\pdfoutput=1

\documentclass[11pt]{article}

\usepackage{naacl2021}
\usepackage[normalem]{ulem}
\useunder{\uline}{\ul}{}
\usepackage{times}
\usepackage{latexsym}
\usepackage[ruled,vlined]{algorithm2e}

\usepackage[utf8]{inputenc}

\usepackage{microtype}
\usepackage{adjustbox}
\usepackage{graphicx}
\usepackage{makecell} %
\usepackage{amssymb}
\usepackage{amsmath}
\usepackage{ctable}
\usepackage{tabularx}
\usepackage{multirow}
\usepackage{tablefootnote}
\usepackage[flushleft]{threeparttable}
\title{Efficient Retrieval Optimized Multi-task Learning}

\author{Hengxin Fun, Sunil Gandhi, Sujith Ravi \thanks{Work done at Amazon Alexa} \\
        Amazon Alexa \\
        \texttt{\{hengxin,gansuni\}@amazon.com, sravi@sravi.org }}

\begin{document}
\maketitle

\begin{abstract}
Recently, there have been significant advances in neural methods for tackling knowledge-intensive tasks such as open domain question answering (QA). These advances are fueled by combining large pre-trained language models with learnable retrieval of documents. Majority of these models use separate encoders for learning query representation, passage representation for the retriever and an additional encoder for the downstream task. Using separate encoders for each stage/task occupies a lot of memory and makes it difficult to scale to a large number of tasks.  In this paper, we propose a novel {\it Retrieval Optimized Multi-task} (ROM) framework for jointly training self-supervised tasks, knowledge retrieval, and extractive question answering. Our ROM approach presents a {\it unified} and generalizable framework that enables scaling efficiently to multiple tasks, varying levels of supervision, and optimization choices such as different learning schedules without changing the model architecture. It also provides the flexibility of changing the encoders without changing the architecture of the system. Using our framework, we achieve comparable or better performance than recent methods on QA, while drastically reducing the number of parameters. 
\end{abstract}

\section{Introduction}

Open domain question answering (QA) is a task of answering natural language questions using a large collection of text documents. Early systems for question answering consisted of complicated pipelines with multiple components for question understanding, document understanding, retrieval and answer extraction. For example, the approach from \cite{ferrucci2010building} consisted of named entity recognizer, question type classifier, lexical answer type detector, and relation detector for question understanding. Recent models have greatly simplified this pipeline in two ways. First, instead of multiple components, pre-trained language models are used for question and passage understanding. Second, recent QA methods consist of two stages---{\it retriever}, that retrieves documents relevant to the query, and {\it reader}, that extracts the correct answer given the relevant documents \cite{chen2017reading}. 

Typically, a separate encoder is used for the question, passage and reader tasks, each initialized with pre-trained language model \cite{karpukhin2020dense}. Using separate encoders for each task occupies a lot of memory and is not scalable to large number of tasks. In this work, we simplify this pipeline by unifying question, passage and the encoder for the downstream tasks in a multi-task framework.  We show that training these methods jointly drastically reduces the number of parameters while achieving comparable performance to the existing methods.

Over the years, several methods for retrieving documents have been proposed in the literature. Traditionally, retrieval of relevant documents was performed by matching keywords using TF-IDF or BM25 based methods. But, recent methods have shown that dense retrieval using neural representations have outperformed the TF-IDF/BM25 based methods on open domain question answering~\cite{lee2019latent,karpukhin2020dense}. But, these methods vary in the supervisory signal and the method used for training the retriever. ORQA\cite{lee2019latent} and REALM \cite{guu2020realm} pre-train the neural retriever using self supervised methods. \cite{lee2019latent} proposes inverse cloze task (ICT) that extracts a query sentence and predicts the block from which the sentence was extracted. REALM uses masked language model for training the neural retriever. DPR~\cite{karpukhin2020dense} is a widely-used approach which trains a dual-encoder that uses supervised question-passage pairs to train the retriever. Other methods like, RAG~\cite{lewis2020retrieval} and Fusion-in-Decoder~\cite{izacard2020leveraging} are based on DPR and use dual-encoder for retrieval of the documents. Although useful, this approach has the disadvantage of requiring supervised training data,  i.e. question-passage pairs. For practical settings and new domains, there is limited supervised training data available and collection of ground truth data is costly. 

In this work, we solve this problem by jointly training self-supervised and supervised tasks in a multi-task fashion. We propose a single encoder architecture with task-specific projection heads. We train the encoder effectively in a unified framework using a scheduled learning strategy. In our scheduled learning framework, we perform ablation studies that demonstrate the effectiveness of self-supervised, semi-supervised and supervised labels on the retrieval accuracy. We show that our single encoder gets comparable performance to the existing methods while using fewer parameters. 

Overall, the contribution of our paper are as follows:
\begin{enumerate}
    \item We propose a novel {\it Retrieval Optimized Multi-task} (ROM) unified framework for jointly training the self-supervised tasks, knowledge retriever, and extractive question answering using a single encoder. 
    \item We introduce a flexible {\it scheduled learning} strategy for training ROM and show that it can effectively scale to multiple tasks, seamlessly switch between them, vary the level of supervision without changing the system architecture. 
    \item We perform extensive experiments and comparison against existing baseline methods. Our results demonstrate that the multi-task ROM system gets comparable results on end-to-end question answering as well as retrieval tasks, despite using fewer parameters. We also show that jointly training the ROM network with self-supervised and retrieval tasks achieves comparable or better performance on the downstream QA task at a low cost both w.r.t. amount of task-specific supervision and model size.
\end{enumerate}

\section{Related work}

Open domain question answering is an important problem and a wide variety of methods have been proposed for this task in the natural language processing literature. Traditional question answering systems consisted of complicated pipelines with multiple components for question understanding, document understanding, retrieval and answer extraction\cite{ferrucci2010building}. Recently, question answering systems have been greatly simplified by the use of sequence-to-sequence models.

One of the most straightforward neural approaches for open domain question answering is to use pre-trained language models like BERT~\cite{devlin2018bert}, T5~\cite{raffel2019exploring} and BART~\cite{lewis2019bart} and fine-tune it for question answering. It has been shown that these models implicitly store a large amount of knowledge and can generalize to open domain question answering~\cite{petroni2020kilt}. They do so without explicitly accessing the knowledge source. However, using pre-trained language models for question answering has several disadvantages. As knowledge is implicitly stored, it is difficult to access it in an interpretable way. Also, the storage space is limited by the size of the network. Hence, the need to train neural models with continually growing sizes.

Consequently, retrieval based methods for open domain question answering have been introduced in the literature to access knowledge. These methods work in two stages. Given a query, a retriever is used to get the relevant documents, and then reader extracts the correct answer using the retrieved documents. Both retriever and reader have gone through significant improvements. The initial retrieval based methods used keyword matching using TF-IDF or BM25 based methods for getting relevant documents~\cite{chen2017reading}. Recently, transformer based architectures have been proposed for learning dense representations for query and passages, and maximum inner product search (MIPS) \cite{johnson2017billion} is used for retrieval. 

Broadly, the dense representations for query and passages are learnt either using self-supervised pre-training or by training using question-passages pairs in a supervised manner. ORQA~\cite{lee2019latent} and REALM~\cite{guu2020realm} use self-supervised pre-training for learning dense representations for the query and passages. ORQA introduced inverse clozure task (ICT), a self-supervised task that extracts sentence from a passage as a pseudo-query and uses context around it as evidence. They showed the retriever pre-trained using ICT and fine-tuned on question-answer pairs outperformed the keyword matching based retrieval. REALM uses salient masked language model (MLM) for pre-training the retriever. DPR~\cite{karpukhin2020dense} introduces dual-encoder architecture for retrieval. It uses question-passages pairs to learn dense representations in a supervised manner. RAG~\cite{lewis2020retrieval} and Fusion-in-decoder~\cite{izacard2020leveraging} uses DPR retriever with a generative model for the reader to perform open domain question answering. 

\begin{figure}[ht]
\centering
\includegraphics[width=0.51\textwidth]{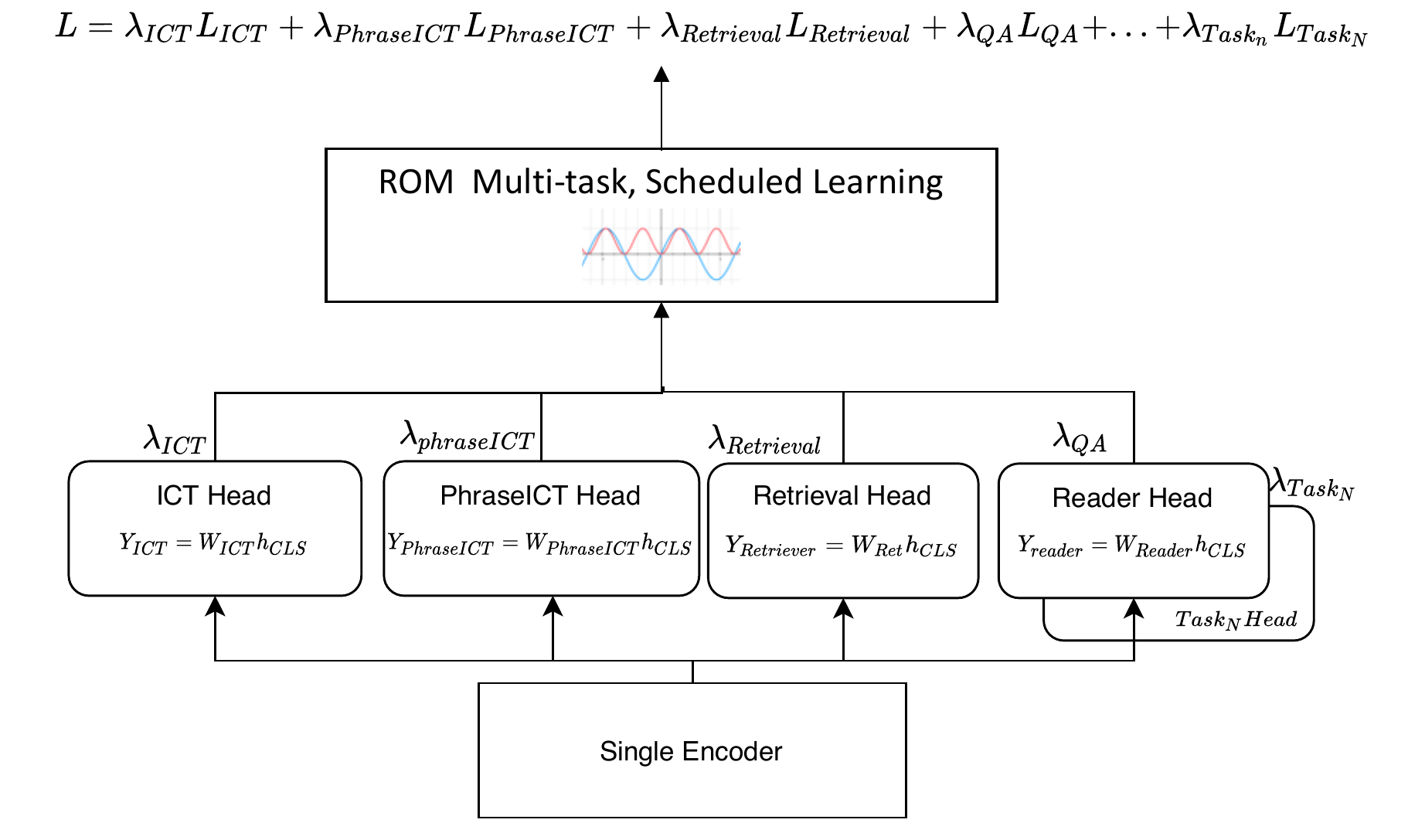}
\caption{Retrieval Optimized Multi-Task (ROM) learning using Single Encoder}
\end{figure}

Although retrieval based approaches outperform the pre-trained language models that store knowledge implicitly in its parameters, they lose the flexibility of using a single model and fine-tuning on multiple downstream tasks. Also, they use a large number of parameters as multiple encoders have to be used during training and inference. For example, DPR uses three BERT encoders, one for learning query embedding, one for passage embedding and another for downstream reading comprehension task. This makes it challenging to perform on-device inference or for resource-constrained settings. In this work, we bridge this gap, by proposing a novel multi-task framework, with retriever, reader and self-supervised tasks implemented using a single encoder and task-specific projection heads. This gives us the flexibility of pre-trained language models, making it easy to scale to large number of tasks while retaining the performance on both retriever and reader. It also enables us to vary levels of supervision and test its impact when training data is limited.

\section{Retrieval Optimized Multi-Task (ROM) Approach}

We propose jointly training the retriever, reader and self-supervised tasks using a single encoder in a multi-task scheduled learning framework (ROM). Learning a common representation space between the retriever and reader tasks on a single model is novel and distinct from end-to-end training schemes or a pipe-lining approach. Although previous approaches have tried to learn the tasks jointly using three separate encoders, we introduce scheduling strategies on a single multi-task model to improve overall convergence of open domain QA tasks on a unified representation space.

\subsection{Multi-Task Learning with Scheduling}

\begin{figure}[t]
\centering
\includegraphics[width=0.5\textwidth]{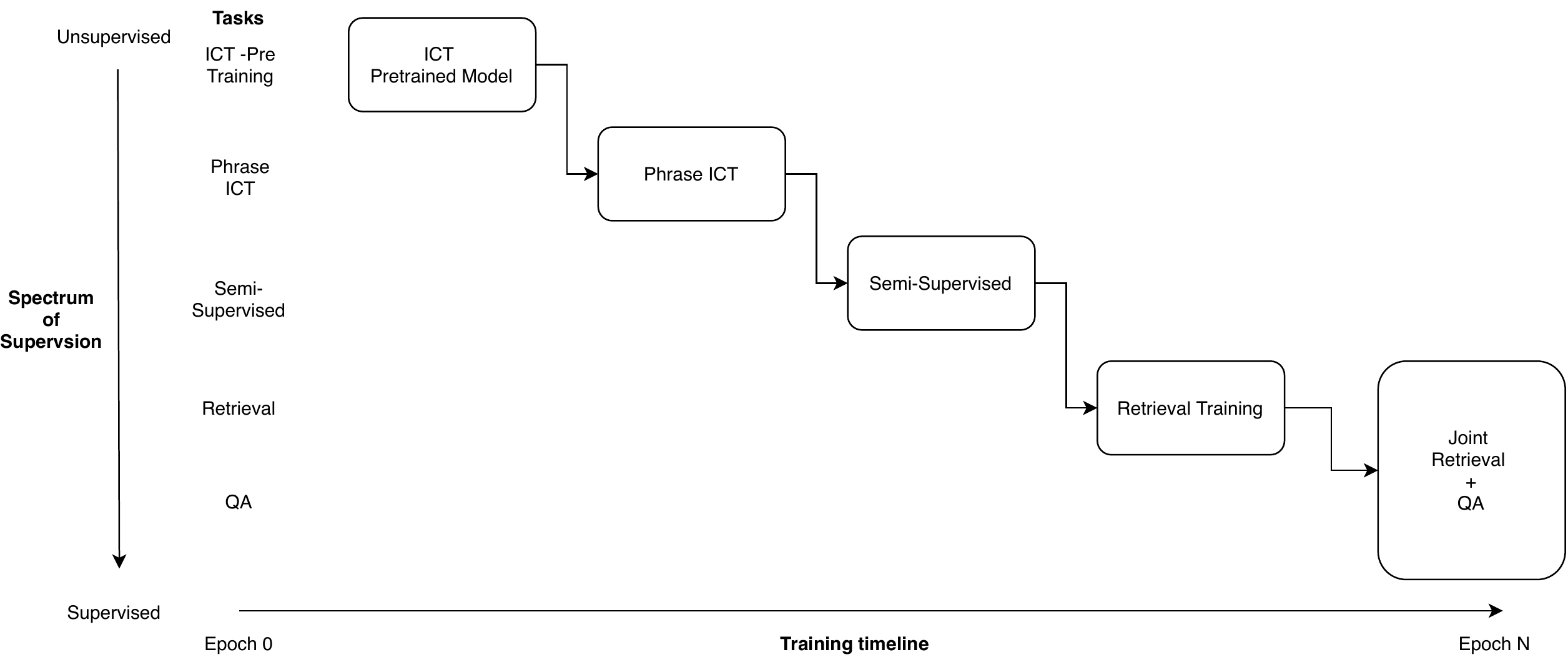}
\caption{Example training timeline for Retrieval Optimized Multi-Task (ROM) learning.}
\label{fig:training_timeline}
\end{figure}

We use BERT-base \cite{devlin2018bert} encoder architecture, but our framework could be easily extended to other pre-trained language models, including encoder-decoder models like BART or T5. Our ROM model is single encoder with task specific projection heads: ICT, Retrieval, and QA--although it can be easily extended to other tasks. For a given task there are task specific losses $L_i$ with corresponding loss weights $\lambda_i$. If losses are of a different scale, they are first normalized to be on the same scale by a normalization factor. Loss weights $\lambda_i^t$ can be adjusted on the fly per epoch $t$, therefore loss weights are assigned based on scheduling function $S(t)$. To further simplify the problem we restrict $\lambda_i$ to be greater than zero and all weights must sum to one. The total loss of a given epoch $t$ is the sum of individual losses weighted by their task specific loss factor.
\begin{align}
&\lambda_i^t :=S(t) \\ \textrm{s.t.} \quad 
\sum_i&\lambda_i =1; \quad \lambda_i\ge0\\
&L^t = \sum_i^n \lambda_i^tL_i
\end{align}

In a typical ROM schedule, we have three phases---pre-training, retrieval and jointly training the retriever and reader. Here, reader refers to the downstream task (QA, in our case).  In the first phase, we pre-train the model with self-supervised ICT and MLM-like tasks, therefore $\lambda_{ICT}$ and $\lambda_{phraseICT}$ are non zero and the other tasks weights are zero. Alternatively, the encoder can also be initialized with an existing pretrained network. In the second phase, we focus on retrieval, i.e., $\lambda_{retrieval}=1$ and all other $\lambda_i=0$. Because QA depends significantly on the quality of retrieval results, retrieval is trained until it converges. Figure~\ref{fig:training_timeline} shows an example timeline for ROM scheduled learning.

\subsection{ROM Joint Optimization}
\label{sec:joint}

Until now we have combined the benefits of pre-training and supervised retrieval. In the final phase, we focus on jointly training the reader and retriever. In contrast to the pipeline approach, which has two models with separate training procedures, we improve both tasks dynamically, similar to end-to-end training but without the complexity. Our method is simpler in that we have a single model that optimizes the tasks jointly on a common representation space.

\begin{algorithm}[h]
\SetAlFnt{\small}
  \scriptsize
\SetAlgoLined
 \While{ Training not finished}{
  retrievalValidation = jointlyTrainRetrieverReader(trainData)\;
  \eIf{ better retrievalValidation}{
  docEmbeddings= computeDocEmbedding(model)\;
  retrievalResults = denseRetrieval(docEmbeddings)\;
  \eIf{ better retrievalResults}{
  trainData = update(retrievalResults)\;
  }{continue}
  }{
  continue\;
  }
 }
 \caption{ROM Joint Optimization}
 \label{algo:RRO}
\end{algorithm}

We introduce a new scheme for updating the retriever and reader dynamically called
\textbf{ROM joint optimization} which is more efficient and less complex than end-to-end training methods and offers more flexibility than pipeline QA systems. In contrast to REALM which requires a constant refreshing of the index, RAG which only updates the query encoder while fixing the document encoder, and DPR which updates the components separately, we refresh our index only when we observe evidence that our retriever has improved--by inspecting retrieval validation. To perform retrieval validation we use average rank validation from \cite{karpukhin2020dense} metric. This metric is less costly to compute and is correlated with good retrieval performance, which makes it more efficient than having to constantly refresh dense embeddings for millions of document.

When refreshing the index, another advantage of having a single model is that common representation space is shared between the query and document encoder. Upon improving the retriever, we generate new document embeddings, create a new index and perform dense retrieval. We only use the new document embeddings and the retrieval results when they have better validation accuracy than previous checkpoints. Once we have better retrieval results, we update the ranking of passages that are used in reader training and resume training of the joint model. Finally, when the reader has new ranking results, the reader must adjust to the new retrieved documents. This interaction between the retrieval and reader tasks can be exploited to improve the convergence of both tasks.

\subsection{Learning schedule for Retriever and Reader in ROM}
\begin{figure}[h]
\centering
\includegraphics[width=0.95\columnwidth]{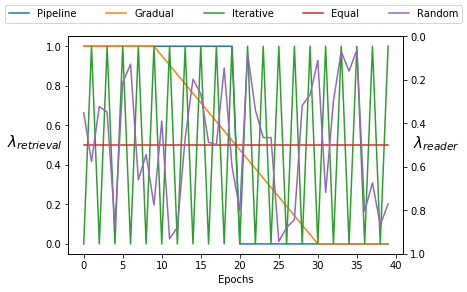}
\caption{ROM schedule for $\lambda_{Retriever},\lambda_{Reader}$}
\label{lambda}
\end{figure}

Open domain QA performance is dependent on both, the retrieval and reader tasks. The tasks weights $\lambda_{retrieval}$ and $\lambda_{reader}=1-\lambda_{retrieval}$, as seen in Figure \ref{lambda}, can be optimized to improve overall performance for a given QA task. We experiment with different weighting schemes across time, such as a {\it pipe-lining}, {\it equal weight}, {\it gradual}, {\it random weight}, and {\it iterative} scheme. In a pipeline schedule, we follow DPR and train the retriever for the first half of the total epochs. Taking the best checkpoint, we produce the retrieval results, which are subsequently used for the last half for the reader---only the reader is trained in the final half. In equal weight regime, $\lambda_{retrieval}=\lambda_{reader}=0.5$. For the iterative scheme, we cycle back and forth between retrieval and reader task, however, $\lambda_{retrieval}$ and $\lambda_{reader}$ are always greater than zero, albeit a very small value. In the random scheme, the weight of $\lambda_{retrieval} \sim  \mathcal{U}(0,1)$, and the  $\lambda_{reader}=1-\lambda_{retrieval}$. Finally, in a gradual schedule the first quarter of training $\lambda_{retrieval}=1$, in the next half there is a linear interpolation to reading task, and in the final quarter $\lambda_{reader}=1$ Next, we describe details of the tasks that are turned {\it on} for optimizing the multi-task ROM model during different training phases.


\subsection{Self-Supervised Tasks}
\subsubsection{Inverse Closure Task (ICT)}
Inverse Cloze Task (ICT) is a pre-training task introduced by \cite{lee2019latent}. We use the same training procedure and loss functions as described in \cite{lee2019latent}. For ICT, we extract sentence as a pseudo question and the context as around the sentence as the evidence. The goal of ICT is to predict the block of evidence from which the pseudo question was extracted. More details on training ICT can be found in \cite{lee2019latent}.

\subsubsection{PhraseICT}
We also test a phrase based version of ICT. We first extract TF-IDF scores for all the words in the wikipedia passages. We select the number of words in the pseudo query, given by $n$. The phrase length, $n$, is randomly selected from 1 to 4. We calculate scores for all spans of length $n$ by summing the TF-IDF scores for the words in the span. We sample the spans as a pseudo query based on their normalized scores. By doing this, we encourage the model to focus on the content words with higher TF-IDF scores. After generating the pseudo queries and corresponding blocks, the rest of the training is the same as ICT pre-training.

\subsubsection{ERNIE 2.0}
ERNIE 2.0 \cite{sun2020ernie} is a transformer language model trained on seven self-supervised tasks. We initialize the ROM model with pretrained ERNIE 2.0 to understand the impact of varying self-supervision strategies on retrieval performance. 

\subsection{Retrieval Task}
The retriever is pre-trained with ICT, PhraseICT or initialized using pretrained ERNIE 2.0 model in phase one. Then in phase two, the retriever is trained in a supervised fashion. We follow the setup of \cite{karpukhin2020dense} closely, however, instead of using two separate models for query and document encoder we use a single BERT model. We use the same projection head to produce $q$ query embedding and $p$ passage embedding. We use the BERT \textbf{[CLS]} hidden representation with the embedding size $d=768$. BERT takes in a query and passages, producing $q=h_{CLS}=\text{BERT}(\text{query})$ and $p=h_{CLS}=\text{BERT}(\text{passage})$, respectively. We use similarity function $sim(q, p)$, computed via an inner product, to score match between query $q$ and a positive ($p^+$) or negative ($p^-$) passage. The retriever is trained using negative log likelihood loss. We refer the reader to DPR \cite{karpukhin2020dense} for more details for training the retriever. 

\subsection{Reader Task (QA)}
Our reader is an extractive model and closely follows DPR. Our reader is not trained on the fly in an end-to-end scheme. Instead we pre-process our reader inputs for a given retrieval ranking prior to training the reader. As the retriever improves, we update the dense embeddings and the retrieved documents and use it to train the reader.  For the reader in ROM, we retrieve 10 passages per query, one positive and the rest negatives.

\section{Experiments}
We evaluate our ROM approach on open domain question answering (QA) benchmarks. In this section, we describe in detail the datasets used, experimental setup and results in both supervised and semi supervised setting. We also vary the learning schedule for training the reader and retriever and understand its effect on the performance.

\subsection{Datasets}
\subsubsection{Wikipedia Data preprocessing}

We use the English Wikipedia dump from Dec. 20, 2018. We reuse the preprocessing script introduced in DPR. This enables us to effectively compare the retrieval and reader scores with the baseline DPR model. The text is extracted from Wikipedia corpus and divided into blocks of 100 words. After preprocessing, the Wikipedia corpus contains 21,015,324 blocks. For training the ROM retriever in self supervised manner we use inverse clozure task(ICT). In ICT, we extract the sentence and use the context around the sentence as the evidence to be predicted. To maintain enough context after removing the sentences, we follow preprocessing strategy from \cite{lee2019latent}  and use the block size of 288 for training the ICT network. On preprocessing, we have 13,353,713 Wikipedia blocks. We use these larger blocks for training the self-supervised tasks. 

\subsubsection{Open Domain Question Answering Datasets}
We perform experiments on two popular question answering benchmarks, TriviaQA~\cite{joshi2017triviaqa} and natural questions~\cite{kwiatkowski2019natural}. Natural Questions consists of Google search queries and answers that are spans from Wikipedia text.  TriviaQA dataset consists of trivia questions with answers scraped from web. DPR~\cite{karpukhin2020dense} discards the questions when they are not able to find matching gold passage due to different version of Wikipedia or preprocessing. To effectively compare our method with DPR, we use the same training, validation and test splits. Detailed statistics of training validation and test splits are shown in Table \ref{tab:dataset_stats}.

\begin{table}
    \centering
    \resizebox{\linewidth}{!}{%
    \begin{tabular}{|l|l|l|l|l|}
    \hline
     Dataset           & \multicolumn{2}{|c|}{Train}     &  Dev   &  Test  \\ \hline
    Natural Questions  &  79168  &  58880  &  8757  &  3610  \\ \hline
    Trivia QA          &  78785   &  60413   &  8837  &  11313 \\ \hline
    \end{tabular}
    }
    \caption{Dataset statistics} 
  \label{tab:dataset_stats}
\end{table}

\begin{table*}[htp]
    \begin{tabular}{|l|c|c|c|}
    \hline
    ~                                              & Type        & \makecell{ Total Parameters \\ (In mn)} &  Exact Match \\ \hline
    ROM network {\it (ours)}                            & Extractive  & 110                      & 40.2                  \\
    HardEM\cite{min2019discrete}                   & Extractive  & 110                      & 28.1                           \\
    PATH Retriever\cite{asai2019learning}          & Extractive  & 110                      & 32.6                           \\
    BM25+DPR\_Reader\cite{lee2019latent}           & Extractive  & 110                      & 32.6                           \\ \hline
    T5 (base-multi-task) \cite{roberts2020much}    & Extractive  & 223                      & 27                             \\
    ORQA\cite{lee2019latent}                       & Extractive  & 330                      & 33.3                           \\
    REALM \cite{guu2020realm}                      & Extractive  & 330                      & 39.2                           \\
    DPR\cite{karpukhin2020dense}                   & Extractive  & 330                      & 40.8\tablefootnote{For equivalent setting of 10 passages per question} /41.4                      \\ \hline
    RAG\cite{lewis2020retrieval}                   & Generative & 620                      & 44                             \\
    Fusion-in-Decoder\cite{izacard2020leveraging}  & Generative & 440                      & 48.5                           \\
    DPR+BM25+T5(large)          & Generative & 1000                     & 51.4                           \\ \hline
    \end{tabular}
        \caption{Comparison of model size (number of parameters) and task performance (exact match scores) for open domain question answering on Natural Questions dataset }
    \label{tab:supervised}
\end{table*}
\subsection{Results}

\begin{table*}[htp]
    \centering
    \resizebox{\linewidth}{!}{
\begin{tabular}{cc|ll|llll|l|}
\cline{3-9}
\multicolumn{1}{l}{}   & \multicolumn{1}{l|}{} & \multicolumn{2}{c|}{Two Models} & \multicolumn{5}{c|}{Single Joint Model}                                      \\ \cline{3-9} 
\multicolumn{1}{l}{}   & \multicolumn{1}{l|}{} & \multicolumn{2}{c|}{Dev Set}    & \multicolumn{4}{c|}{Dev Set}                          & Test Set             \\ \cline{2-9} 
\multicolumn{1}{l|}{} &
  Top K &
  \multicolumn{1}{c}{Pipeline} &
  Random &
  \multicolumn{1}{c}{Equal} &
  \multicolumn{1}{c}{Gradual} &
  \multicolumn{1}{c}{Random} &
  \multicolumn{1}{c|}{Iterative} &
  \multicolumn{1}{c|}{Random} \\ \hline
\multicolumn{1}{|c|}{\multirow{5}{*}{\begin{tabular}[c]{@{}c@{}}Retrieval \\ (Accuracy \%)\end{tabular}}} &
  1 &
  46.37 &
  49.15 &
  51.25 &
  50.42 &
  \textbf{53.68} &
  53.12 &
  53.54 \\
\multicolumn{1}{|c|}{} & 5                     & 69.62      & 71.64     & \textbf{73.88} & 71.66 & 73.10                & 73.07 & 73.49                \\
\multicolumn{1}{|c|}{} & 10                    & 76.06      & 77.31     & \textbf{79.47} & 76.75 & 77.90                & 77.80 & 78.64                \\
\multicolumn{1}{|c|}{} & 20                    & 80.36      & 81.30     & \textbf{83.52} & 80.87 & 81.35                & 81.34 & 82.63                \\
\multicolumn{1}{|c|}{} & 100                   & 86.19      & 86.60     & \textbf{88.56} & 86.40 & 86.62                & 86.41 & 88.33                \\ \hline
\multicolumn{1}{|c|}{\multirow{4}{*}{\begin{tabular}[c]{@{}c@{}}QA   \\ (EM)\end{tabular}}} &
  5 &
  38.30 &
  39.19 &
  39.71 &
  39.00 &
  39.81 &
  \textbf{40.04} &
  39.81 \\
\multicolumn{1}{|c|}{} & 10                    & 39.48      & 40.25     & 40.12          & 39.32 & {\ul \textbf{40.36}} & 39.72 & {\ul \textbf{40.17}} \\
\multicolumn{1}{|c|}{} & 20                    & 39.81      & \textbf{40.33}     & 40.17 & 39.60 & 39.79                & 39.75 & 39.53                \\
\multicolumn{1}{|c|}{} & 100                   & 39.58      & 39.50     & \textbf{39.71} & 39.63 & 39.32                & 39.21 & 39.22                \\ \hline
\end{tabular}}
\caption{Comparison of question answering results for Pipeline versus Scheduled Learning in ROM}
\label{tab:scheduled_table}
\end{table*}

\begin{table*}[]
\resizebox{\linewidth}{!}{%
\begin{tabular}{cc|c|c|c|c|c|c|c|c|}
\cline{3-10}
                     &                                                                         & \multicolumn{4}{c|}{Natural Questions}                            & \multicolumn{4}{c|}{TriviaQA}                                     \\ \cline{3-10}
                     &                                                                         & \multicolumn{2}{c|}{Dev Set}    & \multicolumn{2}{c|}{Test Set}   & \multicolumn{2}{c|}{Dev Set}    & \multicolumn{2}{c|}{Test Set}   \\ \hline
\multicolumn{1}{|c|}{Retrieval}            & \begin{tabular}[c]{@{}c@{}}Labeled dataset \\ (in percent)\end{tabular} & 20             & 100            & 20             & 100            & 20             & 100            & 20             & 100            \\ \specialrule{.1em}{.05em}{.05em}

\multicolumn{1}{|c|}{ROM(pre=ICT)}         & \multirow{3}{*}{Self-supervised}                                        & \textbf{37.51} & \textbf{54.83} & \textbf{43.18} & \textbf{60.88} & \textbf{48.94} & \textbf{67.43} & \textbf{49.1} & \textbf{67.51} \\ 
\multicolumn{1}{|c|}{ROM(pre=PhraseICT)}   &                                                                         & 27.04          & 45.15          & 30.69          & 49.39          & 40.14          & 61.90         & 39.31          & 61.17          \\ 
\multicolumn{1}{|c|}{ROM(pre=ERNIE)}       &                                                                         & 11.11           & 22.15           & 13.77           & 27.01          & 10.00           & 23.73          & 9.71           & 23.45          \\ \specialrule{.1em}{.05em}{.05em}
\multicolumn{1}{|c|}{DPR}                  & \multirow{3}{*}{1}                                                      & 54.06          & 68.39          & 58.01          & 71.99          & 63.29          & 75.38          & 63.07          & 75.74          \\ 
\multicolumn{1}{|c|}{ROM(pre=ERNIE+retriever)} &                                                                         & \textbf{65.36} & \textbf{76.74} & \textbf{67.17} & \textbf{79.25} & \textbf{68.97} & \textbf{79.11} & \textbf{68.38} & \textbf{78.51} \\ \specialrule{.1em}{.05em}{.05em}
\multicolumn{1}{|c|}{DPR}                  & \multirow{3}{*}{5}                                                      & 68.81          & 79.75          & 70.89          & 82.16          & 68.29          & 78.81          & 68.33          & 78.56          \\ 
\multicolumn{1}{|c|}{ROM(pre=ERNIE+retriever)} &                                                                         & \textbf{73.61} & \textbf{82.23} & \textbf{75.24} & \textbf{84.46} & \textbf{73.55} & \textbf{82.11} & \textbf{73.56} & \textbf{81.89} \\ \specialrule{.1em}{.05em}{.05em}
\multicolumn{1}{|c|}{DPR}                  & \multirow{3}{*}{10}                                                     & 70.72          & 80.48          & 71.88          & 83.10          & 72.08          & 80.74          & 71.89          & 80.66          \\ 
\multicolumn{1}{|c|}{ROM(pre=ERNIE+retriever)} &                                                                         & \textbf{73.80} & \textbf{82.68} & \textbf{74.13} & \textbf{83.96} & \textbf{75.53} & \textbf{82.98} & \textbf{75.68} & \textbf{82.91} \\ \specialrule{.1em}{.05em}{.05em}
\multicolumn{1}{|c|}{DPR}                  & \multirow{3}{*}{50}                                                     & 74.50          & 83.21          & 76.51          & 84.74          & 75.30          & 82.54          & 75.40          & 82.77          \\ 
\multicolumn{1}{|c|}{ROM(pre=ERNIE+retriever)} &                                                                         & \textbf{77.12}         & \textbf{84.79}        & \textbf{78.92}        & \textbf{86.32}         & \textbf{77.66} & \textbf{84.26} & \textbf{78.33} & \textbf{84.56} \\ \specialrule{.1em}{.05em}{.05em}
\multicolumn{1}{|c|}{DPR}                  & \multirow{3}{*}{100}                                                    & \textbf{77.83} & 84.83          & 78.4           & 85.4           & 78.99          & 84.78          & 79.4           & 85.0           \\ 
\multicolumn{1}{|c|}{ROM(pre=ERNIE+retriever)} &                                                                         & 77.46          & \textbf{85.03} & \textbf{78.86} & \textbf{86.59} & \textbf{79.71} & \textbf{85.14} & \textbf{80.04} & \textbf{85.45} \\ \specialrule{.1em}{.05em}{.05em}
\end{tabular}
}
\caption{Comparison of retrieval accuracy(top-20, top-100) on question answering datasets for ROM and baseline DPR models with different levels of supervision: {\it self-supervised, semi-supervised and supervised}}
\label{tab:semi_supervised}
\end{table*}

\subsubsection{Open Domain Question Answering}

We evaluate the effectiveness of our method on both retrieval and end-to-end question answering. Following existing work, we use an exact match for end-to-end question answering evaluation and top-K accuracy for evaluating the retrieval. Exact match accuracy is the percentage of questions where predicted answer span matches the ground truth answer. And the top-k retrieval accuracy is defined as a percentage of retrieved passages that contain the answer span. 

We performed all the experiments using 8 NVIDIA Tesla V100 GPUs with 32GB RAM. All the encoders for both baseline DPR and ROM are follow BERT-base-uncased model architecture. We pretrain the ROM model with ICT. Then reader and retriever are trained jointly with random schedule trained for 200 epochs, using learning rate of $1e-5$ and cosine annealing learning rate schedule with warm restarts of every 5 epochs. For training the reader we use a batch size of 6, with 10 passages per question---1 positive and 9 negative passages.  Other hyperparameters follow the default configuration of DPR. 

Table \ref{tab:supervised} shows the exact match scores for open domain question answering on Natural questions dataset. We show that our multi-task ROM framework achieves comparable performance to the DPR despite using one-third of the parameters. We also outperform existing methods with equivalent number of parameters.

\subsubsection{Training with limited labeled data }
In this section, we show the effectiveness of our method on limited training data. We evaluate the retrieval performance for varying levels of supervision, from self-supervised to semi-supervised with $1\%$, $10\%$, $50\%$ of training data to fully supervised ($100\%$). We compare our method to DPR baseline that uses the dual-encoder architecture. 

Both, our single encoder architecture and dual-encoder architecture of DPR models are trained using the same hyperparameters. We trained both the models for 100 epochs with the learning rate of $10^{-5}$ using Adam optimizer and linear scheduling with warm-up and dropout rate of 0.1. We use batch size of 14 and projection embedding size is 768. For training ICT and PhraseICT, we use a similar setup but with wikipedia blocks of size 288. ICT and PhraseICT are trained for 100000 steps and with a learning rate of $10^{-4}$. 

Table \ref{tab:semi_supervised} shows the retrieval performance for self-supervised, semi-supervised and supervised methods. In this work, we explore three self-supervised methods, ICT, PhraseICT and ERNIE 2.0. The retrieval performance is measured using top-k accuracy(Here, $k = 20, 100$ ). Not surprisingly, all supervised models perform better than the three self-supervised models. Interestingly, even $1\%$ of QA task data (i.e. 588 training instances for Natural questions and 604 training instances for triviaQA) can lead to significant improvement in accuracy. Also, our single encoder ROM architecture pre-trained with ERNIE consistently outperforms baseline DPR on both natural questions and TriviaQA dataset. This is despite using half the parameters for training the retriever. 

\subsubsection{Self-Supervised Pretraining}

\begin{table}[]
\resizebox{\linewidth}{!}{%
\begin{tabular}{|c|c|c|c|c|c|c|}
\hline
\begin{tabular}[c]{@{}l@{}}Labeled Data \\ (in percent)\end{tabular} & \multicolumn{2}{c|}{1\%}        & \multicolumn{2}{c|}{10\%}      & \multicolumn{2}{c|}{100\%}      \\ \hline
                                                                     & 20              & 100             & 20             & 100             & 20              & 100             \\ \hline
DPR                                                                  & 54.06          & 68.39          & 70.72         & 80.48          & \textbf{77.83} & 84.83          \\ \hline
ROM(pre=None+retriever)                                              & 54.46          & 68.88          & 70.82         & 80.3           & 77.05          & 84.26          \\ \hline
ROM(pre=ICT+retriever)                                               & 56.64          & 72.44          & 71.07         & 81.09          & 76.36          & 84.48          \\ \hline
ROM(pre=ERNIE+retriever)                                             & \textbf{65.36} & \textbf{76.74} & \textbf{73.8} & \textbf{82.68} & 77.46          & \textbf{85.03} \\ \hline
\end{tabular}
}
\caption{Comparison of retrieval performance with respect to the pre-training methods on development set of Natural Questions dataset  }
\label{tab:pretrain}
\vspace{-4mm}
\end{table}

We also perform ablation study to understand the impact of pre-training method on the retrieval performance. We initialize the ROM model using pre-trained BERT-base-uncased, ICT and ERNIE 2.0 models. These models are then trained with 1\% 10\% and 100\% of the supervised training data. Table \ref{tab:pretrain} shows the results of the ablation study on natural questions development dataset. We can notice that all the ROM models outperform the dual encoder DPR models on $1\%$ of the training data. Also, when training with limited supervised data, initializing ROM using pretrained model leads to significant improvement in retrieval accuracy. Although, the gap reduces as we increase the number of supervised training labels. 

\begin{figure}[h]
\centering
\includegraphics[width=0.90\columnwidth]{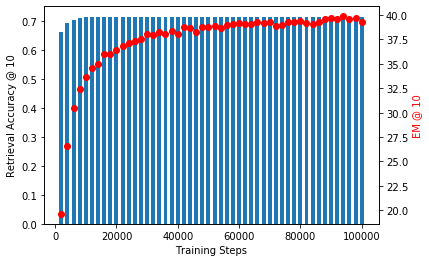}
\caption{ROM Joint Optimization: Interaction of Retriever and Reader (Equal Schedule)}
\label{fig:optimization}
\end{figure}



\subsubsection{Effect of Scheduled Learning}

In this section, we evaluate the impact of scheduling the retriever and reader training on question answering performance. We compare five scheduling methods, pipe-lining, equal weight, random weight, iterative, and gradual scheme. Pipe-lining schedule follows existing methods like DPR, where retriever is trained first and then the reader. Consequently, two checkpoints are used for the inference. The trained retriever is used to generate document and query embedding, and the reader checkpoint is used for the question answering task. This is distinct from other schedules described in Table \ref{tab:scheduled_table} where a single checkpoint with retrieval projection head is used for generating question and document embedding, and reader projection head is used for question answering. 

Table \ref{tab:scheduled_table} shows the retrieval and reader performance for different scheduling mechanisms. We can observe that a unified ROM model with shared parameters consistently outperforms the pipeline approach. Scheduled learning outperforms on both retrieval accuracy as well as exact match question answering accuracy. This suggests that the retriever and reader tasks are complementary and benefit from joint training with a common representation space. We also experimented using the two best checkpoints for scheduled learning--although still trained jointly--and found comparable results to the single model. For example, a model trained on a random schedule with two models performed slightly worse in the downstream QA task with 40.25 as compared to 40.36 in single joint model. 

When comparing different schedules, an equal schedule where the task weights are constant throughout the training are more conducive to improving the retriever. Random and iterative schedule performs marginally better for the downstream QA task. This is surprising because in the gradual schedule, the task weight for the later epochs are focused solely on the QA task. We suspect that the equal, random, and iterative schedule have a regularizing effect on the QA task and thus helps generalization.

For training reader and retriever jointly, we use the optimization algorithm described in section \ref{sec:joint}. In Table \ref{tab:scheduled_table}, we see improvement across the board using ROM joint optimization when compared to Table \ref{tab:semi_supervised}. Figure \ref{fig:optimization} shows the interaction between EM scores and top 10 retrieval accuracy for ROM model with equal task weights schedule. We can observe from Figure \ref{fig:optimization}, that the joint optimization has a significant effect in the early stages of training. This is because in the early epochs the retrieval scores can be easily improved upon, thus initiating the generation of new embeddings. Throughout the training, the dense embeddings were generated five times due to improvement in the validation retrieval scores. 

\section{Conclusion and Future work}
In this work, we propose a novel multi-task framework (ROM) for efficiently training self-supervised tasks, retriever and extractive question answering. We show that this enables us to easily switch between multiple tasks, vary levels of supervision without changing the system architecture. We show that our method outperforms existing methods on open-domain question answering with lower supervision and fewer parameters. In future, we plan to extend this framework to additional knowledge intensive tasks like fact checking, summarization and dialog generation.

\bibliography{anthology,custom}
\bibliographystyle{acl_natbib}


\end{document}